\documentclass{article} 
\usepackage{iclr2025_conference,times}

\usepackage{microtype}
\usepackage{graphicx}
\usepackage{subfigure}
\usepackage{booktabs} 

\usepackage{amsmath,amsfonts,bm}

\def\eqref#1{equation~\ref{#1}}

\def\1{\bm{1}}

\DeclareMathAlphabet{\mathsfit}{\encodingdefault}{\sfdefault}{m}{sl}
\SetMathAlphabet{\mathsfit}{bold}{\encodingdefault}{\sfdefault}{bx}{n}

\usepackage{hyperref}
\usepackage{url}
\usepackage{multirow}
\usepackage{makecell}
\usepackage{algorithm}
\usepackage{algorithmic}

\usepackage{amsmath}
\usepackage{amssymb}
\usepackage{mathtools}
\usepackage{amsthm}

\usepackage{xcolor}

\usepackage{listings}

\newcommand{\mname}[0]{ProgGen}

\usepackage[capitalize,noabbrev]{cleveref}

\usepackage[textsize=tiny]{todonotes}

\title{Programmatic Video Prediction Using Large Language Models}

\author{Hao Tang\thanks{Work done during an internship at MERL.}~\quad Kevin Ellis \\
Department of Computer Science\\
Cornell University\\
Ithaca, NY, USA \\
\texttt{\{ht383,kellis\}@cornell.edu} \\
\And
Suhas Lohit \quad  Michael J. Jones \quad  Moitreya Chatterjee \\
Mitsubishi Electric Research Laboratories (MERL) \\
Cambridge, MA, USA \\
\texttt{\{slohit,mjones,chatterjee\}@merl.com} 
}

\iclrfinalcopy

\begin{document}



\maketitle

\newcommand\ie{\textit{i.e.}}
\newcommand\etc{\textit{etc.}}
\newcommand\name{ProgGen}
\newcommand{\etal}{\textit{et al.}}

\begin{abstract}
The task of estimating the world model by describing the dynamics of a real world process assumes immense importance for anticipating and preparing for future outcomes and finds wide-spread use in applications such as video surveillance, robotics, autonomous driving, \etc~ This task entails synthesizing plausible visual futures, given a few frames of a video -- necessary to set the visual context for the synthesis. Towards this end, different from end-to-end deep learning based approaches for video frame prediction, we propose \name{} -- which undertakes the task of video frame prediction by synthesizing computer programs which represent the dynamics of the video using a set of neuro-symbolic, human-interpretable set of states (one per frame) by leveraging the inductive biases of Large (Vision) Language Models (LLM/VLM). In particular, \name{} utilizes LLM/VLM to synthesize computer programs to: (i) estimate the states of the video, given the visual context (\ie the frames); (ii) predict the states corresponding to future time steps by estimating the transition dynamics; (iii) render the predicted states as visual RGB-frames. Empirical evaluations reveal that our proposed method outperforms competing techniques at the task of video frame prediction in two challenging environments: (i) PhyWorld and (ii) Cart Pole. Additionally, \name{} permits counter-factual reasoning and editability, attesting to its effectiveness and generalizability. 
\end{abstract}    
\section{Introduction}
\label{sec:intro}



The task of extrapolating the dynamics of a real world process into the future is an essential one and necessitates an understanding of the world model governing the dynamics of the process. As artificial intelligence systems become ubiquitous and get deployed to work side-by-side with humans, for tasks such as video surveillance~\cite{wang2018background}, autonomous driving~\cite{jain2015car}, robotics~\cite{cherian2024llmphy}, it becomes increasingly more important that they be capable of reasoning about the dynamics of real world phenomena, given a visual context, and synthesize plausible futures. 

With the recent success of deep neural networks, the research community has resorted to deploying such models for the task of video frame prediction~\cite{babaeizadeh2017stochastic,denton2018stochastic,chatterjee2021hierarchical}. Models developed for this task are capable of generating the future frames of a video, given a few past frames -- in order to set the visual context for the generation. Lately, diffusion models~\cite{blattmann2023stable,hoppe2022diffusion} have shown the most promise for this task. These models have an encoder-decoder architecture. In the encoding stage, a small magnitude of noise (usually sampled from a Gaussian distribution) is iteratively added to a sample drawn from the training data, potentially until the real data is indistinguishable from pure (Gaussian) noise. The goal of the decoding phase then, is to iteratively de-noise the sample to eventually recover the original data. Despite their success, the widespread adoption of such approaches is limited primarily by their reliance on copious amounts of training data and their lack of interpretability. Moreover, such models are only effective when the data on which the inference is to be performed is largely similar to the one on which they have been trained.

Existing physics-based approaches~\cite{wu2015galileo,liu2024physgen} for the task of video frame generation/prediction, seek to estimate physical properties of the objects that are present in the video, such as their mass, position, \etc~ either using perception modules or by having users provide initialization of those attributes. However, such approaches either pre-define these attributes or pre-define the world model governing the dynamics of the objects in the video, limiting their applicability.

\begin{figure*}[t]
    \centering
    \includegraphics[width=0.9\textwidth]{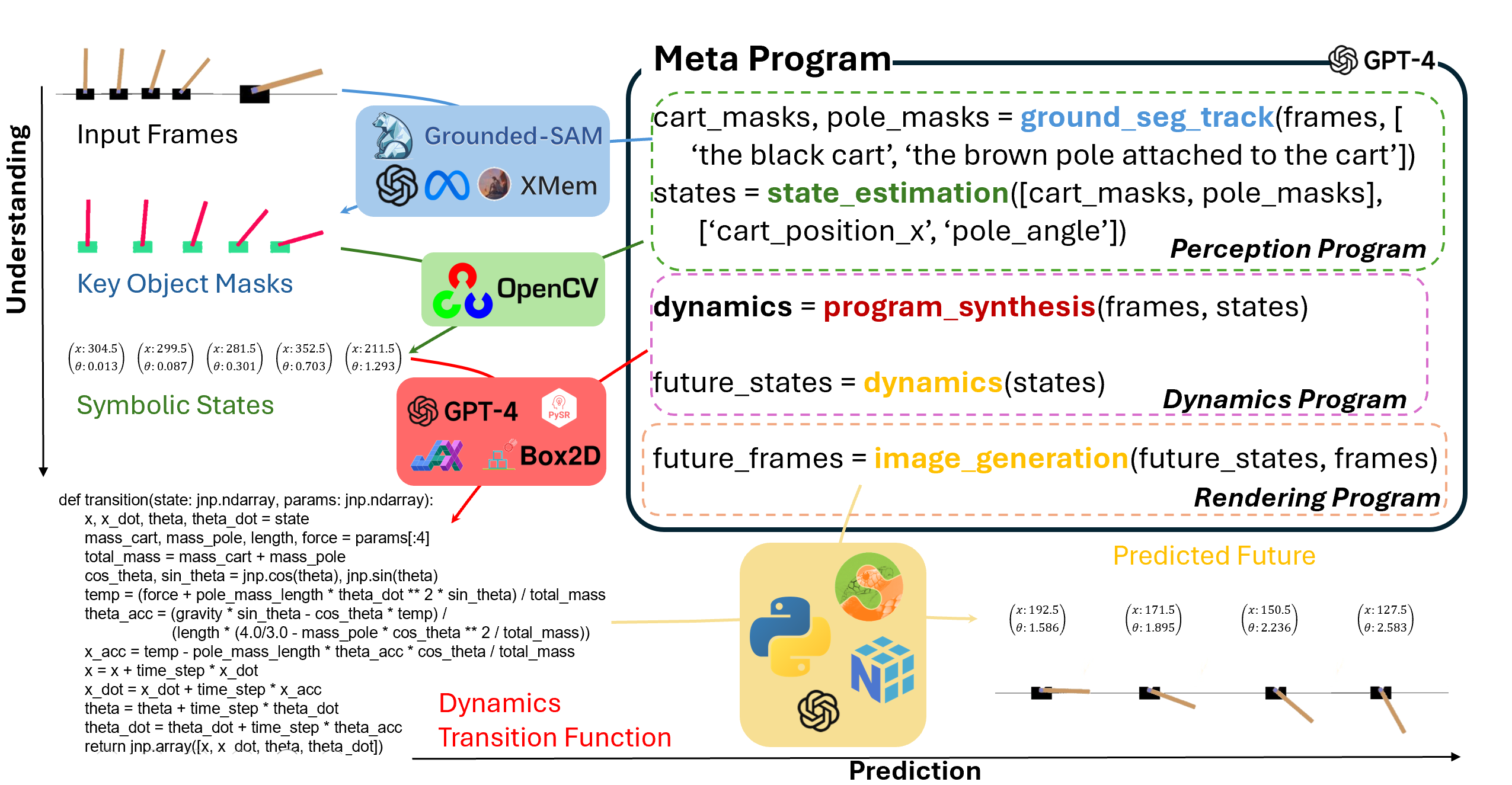}
    \caption{Overview of \name{} showing the meta program that solves the video prediction task by: (i) generating a program for perception and estimation of object states ($\mathcal{P}$), (ii) estimating the world model (\ie the dynamics of the video) program synthetically ($\mathcal{D}$), and (iii) rendering future frames using a rendering program ($\mathcal{R}$). The figure also illustrates the output of these steps for the Cart Pole setting, including the principal objects that have been discovered (such as ``the black cart''), their states, the generated program for estimating the dynamics and the predicted frames of the video.} 
    \label{fig:overview}
    \vspace{-0.6cm}
\end{figure*}

In this work, we mitigate the aforementioned challenges by developing a novel, physics-grounded, neuro-symbolic, program synthesis-based video prediction model that leverages the inductive biases of Large Language/Vision-Language Models (LLM/VLM) 
for improved sample efficiency, generalizability, and interpretability. Additionally, our formulation permits user-guided editability and counter-factual reasoning on the video frames which are to be generated as well. At its core, our proposed method called \name{}, represents the dynamics of a video using a set of neuro-symbolic states, each defined by a set of human-interpretable, physics-grounded attributes, such as position or velocity of an object in the video. To enable such a representation, \name{} employs a novel, programmatic perception pipeline (denoted by $\mathcal{P}$) to estimate these states from the visual context of a video and another program (denoted by $\mathcal{D}$) to learn the dynamics, so as to be able to extend the state estimates into the future, using a VLM. A programmatic rendering pipeline (denoted by $\mathcal{R}$), also generated by a VLM, then enables \name{} to decode these predicted states into visually plausible RGB-frames. We leverage pre-trained deep-neural segmentation and tracking models (such as Grounded-Segment Anything~\citep{ren2024grounded}) to facilitate the perception process needed to estimate the states and adopt a symbolic, physical simulator~\citep{feronato2012box2d} to simulate the objects' dynamics in the pixel space. Moreover, the human-interpretable nature of the estimated latent states, lends them to human manipulation allowing for editing or counter-factual reasoning on the videos. Figure~\ref{fig:overview} presents an overview of the \name{} pipeline. 

In order to empirically validate the efficacy of \name{}, we present empirical evaluations on synthetic environments -- PhyWorld~\cite{kang2024far} and Cart Pole~\cite{gym}. Aided by the inductive biases of the LLM/VLM, our method learns a customized video prediction model with no more than 10 training samples and performs competitively to the state-of-the-art diffusion models that are trained with upto three million samples. 
Moreover, our model generalizes much better to out-of-distribution inputs than the diffusion models can. 

In summary, the contributions of our work, are as follows:
\begin{itemize}
    \item We develop a novel, programmatic video prediction model, called \name{}, leveraging the inductive biases of LLM/VLM in order to do away with the need for large-scale training data.
    \item \name{} undertakes the video prediction task by representing the dynamics of the video using a set of neuro-symbolic, human-interpretable set of states.
    \item Empirical evaluations reveal that \name{} outperforms competing approaches at the task of video frame prediction in two challenging environments: (i) PhyWorld (ii) Cart Pole.  
    \item We show that \name{} permits counter-factual reasoning, editability, and interpretable video generation attesting to its effectiveness and generalizability for video prediction tasks.
    
\end{itemize}

\section{Related Works}
\label{sec:related}

\paragraph{End-to-end Models for Video Prediction:}
Recent years have seen an explosion of works that use end-to-end deep learning-based methods for video frame prediction in unconstrained environments, with or without text prompts, trained on datasets with thousands of videos~\cite{lu2023vdt,nikankin2022sinfusion,sun2023moso,chatterjee2021hierarchical,yu2023video,he2022latent,ho2022video}. Most of these models adopt a diffusion model-based pipeline. 
Some prominent examples include methods like Stable Video Diffusion~\cite{blattmann2023stable}, RaMViD~\cite{hoppe2022diffusion}, and TI2V-Zero~\cite{ni2024ti2v} that allow feeding a few initial frames of the video to the diffusion model in order to generate plausible future frames of the videos. These models currently tend to generate very short videos ($\sim1$ minute long), before losing temporal consistency and do not allow for easy and interpretable control over the video generation process. Furthermore, they do not implicitly learn the physics of the world accurately, as shown by Kang \etal~\cite{kang2024far} where the diffusion models are unable to learn world models even in very simple settings. Our proposed method attempts to address these challenges by adopting a program synthetic approach for the task, representing the dynamics of the video by a set of neuro-symbolic, human interpretable states. 

\paragraph{Program Generation and Physics-based Approaches for Video Prediction:} 
Unlike end-to-end methods, program generation and physics-based methods adopt a neuro-symbolic approach in order to learn the dynamics of a system.
 In the case of Galileo~\cite{wu2015galileo}, a Markov-Chain Monte Carlo (MCMC)-based sampling method is adopted to infer a set of physical properties of the objects in the video, such as mass, position, friction, \etc~ A 3D physics engine is then used to simulate the future frames of a video using these estimated quantities. Differently, our proposed method, uses a program synthetic approach to infer the physical properties by leveraging LLM/VLMs.   
DreamCoder~\cite{ellis2023dreamcoder} uses a wake-sleep algorithm to synthesize programs from a small set of input examples and has been shown to learn natural concepts that are applicable in a variety of applications including drawing simple graphics. Similar to our work, WorldCoder~\cite{tang2024worldcoder} uses an LLM to generate programs that can be used to simulate the environment. However, these approaches only work with environments with discrete states and is not directly applicable to cases like ours where continuous parameters (like object properties) have to be estimated. PhysGen~\cite{liu2024physgen} takes an image and attributes (like force/torque) as input, to generate the dynamics of objects by using a strong perception module and a physics engine. However, this method only deals with rigid bodies and the world model is not learned by observing videos, unlike in our work. 
Concurrent to our work, LLMPhy~\cite{cherian2024llmphy} was proposed which also uses an LLM to generate code in order to predict the physical properties of the objects in a video. However, what these properties are, is pre-determined and so is the nature of motion (such as quadratic, linear, \etc), while in our case both the attributes and the dynamics are automatically determined.
Symbolic reasoning (SR) methods like PySR~\cite{cranmer2023interpretable} that use evolutionary optimization and LLM-SR~\cite{shojaee2024llm} learn dynamical equations from data. However, unlike these approaches, \name{} provides a complete, program-synthetic pipeline that incorporates perception, dynamics, and generation into a single interpretable system.
\section{Preliminaries}
\label{sec:prelim}



A video $V = \{f_0, f_1, f_2, \dots, f_T\}$ with $T+1$ frames, 
can be thought of as a collection of $T+1$ random variables $f_i$, each denoting a RGB frame of the video, \ie  $f_i \in \mathbb{R}^{H \times W \times 3}$, where $H$ and $W$ denote the height and width of each frame. Given a set of $F+1$ initial frames of the video ($0 \leq F < T$), to set the visual context, the task of video frame prediction entails forecasting the remaining frames, $f_{F+1}$ onwards through $f_T$. Addressing this task requires designing a model $p_\theta$, parameterized by a set of parameters $\theta$, which takes as input the first $F+1$ frames of an unseen video and predicts the remaining frames. This can be represented as:
\[
\{f_i\}_{i=F+1}^T = p_\theta(\{f_j\}_{j=0}^F; \{\psi_j\}_{j=0}^F),
\]
where $\psi_j$ denotes a possible latent state of the model (one for every predicted frame), in case its design leverages such a formulation (such as in the case of recurrent neural models).

\section{Proposed Method}
Typically, the aforementioned predictive models are instantiated via parameter-intensive deep neural network-based models~\cite{blattmann2023stable,ni2024ti2v}, which are plagued by two key challenges: (i) such models often require large volumes of in-domain training data 
and (ii) the lack of interpretability of the prediction process. 
In order to address these key challenges, in this work, we adopt a physics-grounded, neuro-symbolic approach for the task of video frame prediction. In broad strokes, we train our proposed approach called \name{}, to estimate a set of neuro-symbolic states (each with its own physics-grounded, human-interpretable attributes such as position of an object, its velocity, \etc), which best explains the phenomena observed in frames $f_{F+1}$ through $f_T$, while maintaining consistency with the frames $f_{0}$ through $f_F$. 
This process is realized by estimating three Python programs, derived from a large Vision-Language Model (VLM), like GPT-4~\cite{achiam2023gpt}. The first of these is called the \emph{Perception Program} ($\mathcal{P}$), which maps any RGB-frame $\{f_i\}_{i=0}^T$ to a state ($\{s_i\}_{i=0}^T$). A state is a collection of physics-grounded, human-interpretable attributes, such as position of an object, angle between an object's parts, \etc~ $\mathcal{P}$ automatically determines which attributes are the most applicable for a given video. The second program, called the \emph{Dynamics Estimation Program} ($\mathcal{D}$), 
seeks to forecast future physics-grounded states given a past set of states. The final program, called the \emph{Rendering Program} ($\mathcal{R}$), maps these states ($\{s_i\}_{i=0}^T$) back to the RGB-frames. The inference pipeline of \name{} operates by having $\mathcal{P}$ derive the state-representation of the seen frames in a video ($\{f_i\}_{i=0}^F$), followed by $\mathcal{D}$ forecasting the set of neuro-symbolic states corresponding to the unseen frames ($\{s_i\}_{i=F+1}^T$), given the state-representation of the seen frames in the video ($\{s_i\}_{i=0}^F$). Finally $\mathcal{R}$ decodes the predicted states, corresponding to the unseen frames, in order to synthesize the predictions for the corresponding RGB-frames ($\{\hat{f}_i\}_{i=F+1}^T$). 
Algorithm~\ref{alg:proggen-gen} presents an outline of the inference regime of \name{} while Figure~\ref{fig:overview} presents an overview. 

\subsection{Neuro-Symbolic Video Understanding and Prediction}

At a high-level \name{}, subsumes two tasks: (i) \emph{Video Understanding:} Entails estimating the neuro-symbolic states, corresponding to the seen frames, $f_0$ through $f_F$ of a video, using $\mathcal{P}$, such that when these states are decoded into RGB-frames, using $\mathcal{R}$, they maximize the likelihood of the seen frames. 
(ii) \emph{Video Prediction:} Wherein, $\mathcal{D}$ and a handful of trainable parameters $\theta$, are used to learn the video dynamics and forecast the states corresponding to the unseen frames, $f_{F+1}$ through $f_T$ of a video. This is followed by the use of $\mathcal{R}$ to decode these states to RGB-frames. During training, both the video understanding and video prediction steps can leverage all $T+1$ frames of a video in the training set to determine the $\mathcal{P}$, $\mathcal{D}$, $\theta$, and $\mathcal{R}$ which best explain the said videos. For inference on any video, rather than directly predicting the unseen frames, $f_{F+1:T}$, given the seen frames $f_{0:F}$, as is common in conventional encoder-decoder based frame prediction approaches~\cite{denton2018stochastic,chatterjee2021hierarchical}, we instead use $\mathcal{P}$ to first encode the seen frames into states, and then use $\mathcal{D}$, $\theta$, and $\mathcal{R}$, to predict and render the unseen frames. Mathematically, the perception step can be represented by the following notation:
\[
s_i \sim p\left(\cdot | f_{i}; \mathcal{P}\right); \forall i \in \{0, \dots, F\}
\]
While the prediction step can be described, using the following two-step procedure:
\[
s_i \sim h\left(\cdot | s_{i-1}; \mathcal{D}, \theta\right); \forall i \in \{F+1, \dots, T\}
\]
\[
f_i \sim r\left(\cdot|s_i; \mathcal{R}\right); \forall i \in \{F+1, \dots, T\}
\]

\subsection{Programs and States}
Unlike conventional encoder-decoder based approaches~\cite{ni2024ti2v,blattmann2023stable}, the latent states ($s_i$) in \name{} are symbolic, human-interpretable, and carry semantic meaning. In particular, they are described by physics-based, human-interpretable attributes, such as \emph{position} of an object, its \emph{velocity}, \etc~ The design of \name{}, allows for $\mathcal{P}$ to figure out what attributes might be the most fitting to represent the states. Such a capability allows for much more flexibility of representation, thereby enhancing the generalizability of \name{}. $\mathcal{P}$ uses state-of-the-art visual reasoning tools like Grounded-SAM~\cite{ren2024grounded} for detecting and localizing the objects in the video and XMem~\cite{cheng2022xmem} for ensuring that the detections are consistent over the frames. Likewise, $\mathcal{D}$ and $\theta$ allow for flexibility in representing the dynamics of the video, with $\theta$ capturing the global parameters in the environment, such as the gravity. 

\begin{figure}[t]
    \centering
    \includegraphics[width=0.6\linewidth]{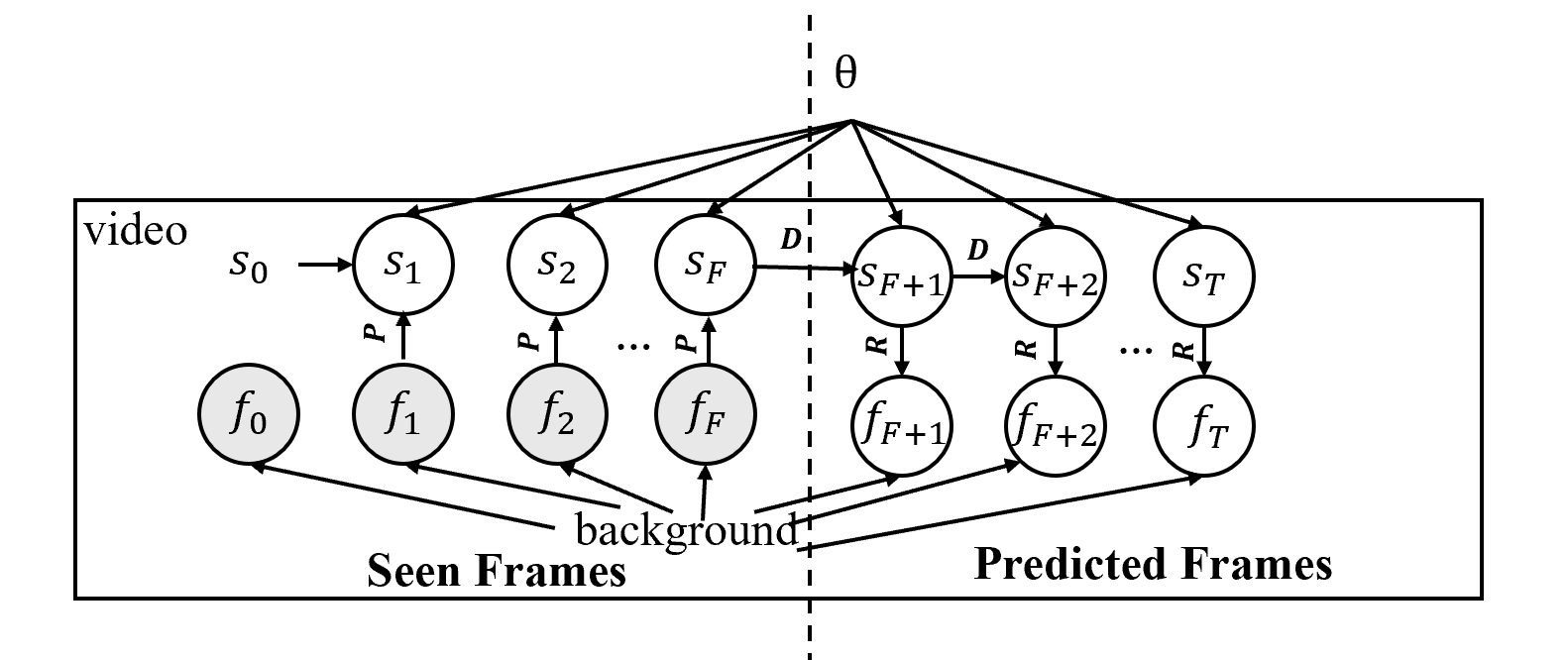}
    \caption{Plate diagram depicting the graphical model of \name{}, during inference. The first $F+1$ frames are used to set the visual conditioning while the subsequent frames are predicted. }
    \label{fig:pgm}
    \vspace{-0.7cm}
\end{figure}

Figure~\ref{fig:pgm} presents a plate diagram of our proposed framework. Each frame $f_i$ of a video is derived from its underlying latent state $s_i$, which in turn is influenced by the previous states $s_{0:i-1}$. For completeness, we also assume a random variable, $background$, to be capturing the details of the static background environment of the video.\footnote{For the sake of brevity, we ignore this random variable, which can be introduced into the formulation without loss of generalizability.} 
The program $\mathcal{P}$ captures a disentangled structure of the scene in a video, such as how many objects there are, if two objects are connected, if an object is dynamic or static, \etc~
The dynamics of a video are captured by the evolution of the states, following the rules specified by the program, $\mathcal{D}$ and the global parameters, $\theta$. 
In particular, $\mathcal{D}$ lays out the laws of physics that principally explain the dynamics in the video, such as laws of inertia, laws of conservation of energy and momentum \etc~ 
The program $\mathcal{R}$ is responsible for mapping the states of the predicted frames to the RGB-domain, using a physics-based simulator. 

We seek to leverage the effective program generation capacity of large Language/Large Vision-Language Models (LLM/VLM)~\cite{achiam2023gpt} to propose hypotheses for $\mathcal{P}$, $\mathcal{D}$, and $\mathcal{R}$, given the frames of a training video, $f_{0:T}$. To aid the VLM in this search, we design
Affordance Rules (ARLs) which describe certain constraints governing the dynamics of the environment to aid the simulation of the dynamics in the pixel space.
For instance, the ARL for a rigid body motion environment setting could be that affine transforms capture the motion of objects in the RGB-frame. 
This significantly reduces the number of candidate hypotheses programs which fit the training data, thereby making the searching task by the LLM/VLM, more efficient.  

\subsection{Training \name}
\label{sec:learnprog}

Training \name{} entails 
estimating the programs $\mathcal{P}$, $\mathcal{D}$, and $\mathcal{R}$, as well as a handful of global parameters $\theta$ in order to maximize the likelihood of the frames in each of the $N$ training videos. For each video, this can be captured by the following probabilistic formulation: 
\begin{align*}
    \mathcal{L}(f_{0:T}; \mathcal{P}, \mathcal{D}, \theta, \mathcal{R}) :=  p(f_{0:T}|s_0; \mathcal{P}, \mathcal{D}, \theta, \mathcal{R}) \\
    = \int_{s_{0:T}}p(f_{T}|f_{0:T-1}, s_0; \mathcal{P}, \mathcal{D}, \theta, \mathcal{R}) p(f_{0:T-1}|s_0; \mathcal{P}, \mathcal{D}, \theta, \mathcal{R}) \text{d} s_t \\
\end{align*}
\begin{align*}
    = \int_{s_{0:T}} p(f_{t}|s_{t}) p(s_{t}|s_{t-1}) p(s_{t-1}|f_{0:t-1}) p(f_{0:t-1}|s_0; \mathcal{P}, \mathcal{D}, \theta, \mathcal{R})\text{d}s_t \\
    = \int_{s_{1:T}} \prod_{t=1}^{T} r(f_{t}|s_{t}; \mathcal{R}) h(s_{t}|s_{t-1}; \mathcal{D}, \theta) p(s_{t-1}|f_{t-1}; \mathcal{P}) \text{d} s_t \\
\end{align*}

The training objective may then be represented as follows:


\begin{align}
\mathcal{P}, \mathcal{D}, \theta, \mathcal{R} =\arg\max_{\widetilde{\mathcal{P}}, \widetilde{\mathcal{D}}, \widetilde{\theta}, \widetilde{\mathcal{R}}} \mathcal{L}(f_{0:T}; \widetilde{\mathcal{P}}, \widetilde{\mathcal{D}}, \widetilde{\theta}, \widetilde{\mathcal{R}})
\label{eq:likelihood}
\end{align}



In order to aid interpretability and for ease of design, we 
assume $r(f_t|s_t; \mathcal{R}) \sim \mathcal{N}(f_t, \Sigma),$ where $\Sigma$ is an isotropic constant. Since maximizing the likelihood is equivalent to maximizing its logarithmic variant, this results in the following optimization setup, over the training set:
\begin{align}
\log \mathcal{L}(f_{0:T}; \mathcal{P}, \mathcal{D}, \theta, \mathcal{R}) \propto \sum_{i=1}^N \sum_{t=0}^T ||\hat{f}_t^i - f_t^i ||^2,
\label{eq:log_likelihood}
\end{align}
where $\hat{f_t^i}$ denotes the estimate of the $t^{th}$-frame by \name{} of the $i^{th}$ video while $f_t$ denotes the true frame in the training video.


\paragraph{Two-stage Training Optimization:}

Foundation models such as large language models (LLMs) or visual language models (VLMs) have shown impressive abilities to synthesize programs~\cite{tang2024worldcoder} and analyze the discrete structures of the scene~\cite{ren2024grounded}. However, they are not as effective in estimating continuous parameters~\cite{tang2024worldcoder}. We thus introduce a two-stage pipeline to train \name{}. In the first stage, the programs, $\mathcal{P}$, $\mathcal{D}$, and $\mathcal{R}$ are estimated, while the trainable continuous parameters $\theta$ are estimated in the next. To enable the learning of the continuous parameters, the LLMs are prompted to synthesize programs with placeholders for the continuous parameters, $\theta$:
$prog\sim q_\text{LLM}(\cdot|f_{0:T})$.
The continuous parameters are then learned by maximizing the likelihood of the seen frames of a given video, as shown in Eq.~\ref{eq:likelihood}. For non-differentiable programs, we use black-box optimizers such as the Powell's method~\cite{powell2006fast} to learn the parameters. If the Affordance Rules requires LLMs/VLMs to generate differentiable programs, e.g., using JAX, we then adopt higher-order optimizers such as L-BFGS~\cite{fei2014parallel} for more efficient optimization. These optimizations are executed till convergence or until a maximum number of iterations, whichever occurs earlier.

\paragraph{Surrogate Loss for Efficient Optimization:}
The typical pixel-level training loss, shown in Eq.~\ref{eq:log_likelihood}, while being differentiable and easy to optimize is not sensitive to the accuracy of the reconstructed frames. For instance, if in a scene a large, white background occupies most of the pixels, Eq.~\ref{eq:log_likelihood} yields a low loss even if a completely blank frame is predicted. Moreover, the computational cost to run $\mathcal{R}$ for every frame in the training video is expensive. As a fix to this, we independently try to achieve estimates for the attributes listed by \name{}, for defining the state of the video. Such attributes could include positions of objects, angles, \etc~ and could be independently estimated using tools such as Segment Anything Model (SAM)~\cite{ren2024grounded} and appropriate functions from OpenCV~\cite{bradski2000opencv}. 
Along the lines of Eq.~\ref{eq:log_likelihood}, the likelihood estimation task for each training set video, then becomes:


\begin{align}
\mathcal{L}(s_{0:T}; \mathcal{P}, \mathcal{D}, \theta) :=  q(s_{0:T}|f_{0:T}; \mathcal{P}, \mathcal{D}, \theta) \\
\propto \sum_{t=0}^T ||\hat{s_t} - s_t ||^2,
\label{eq:inv_log_likelihood}
\end{align}
where $\hat{s_t}$ denotes the estimate of the $t^{th}$-state by \name{} while $s_t$ denotes the true state value. 

\begin{algorithm}[h]
\caption{Inference with \name{}}
\label{alg:proggen-gen}
\begin{algorithmic}[1]
\REQUIRE Video $f_{0:F}$, $s_0$, $\mathcal{P}$, $\mathcal{D}$, $\theta$, $\mathcal{R}$
\ENSURE Generated video $f_{F+1:T}$
\FOR{$t=1$ to $F$}
\STATE $s_t\gets \mathcal{P}(f_{t},\theta)$
\ENDFOR
\FOR{$t=F+1$ to $T$}
\STATE $s_t\gets \mathcal{D}(s_{t-1}),\theta$
\STATE $f_t\gets \mathcal{R}(s_{t})$
\ENDFOR
\STATE \textbf{return} $f_{F+1:T}$
\end{algorithmic}
\end{algorithm}

\section{Experiments} 
\label{sec:exp}

\subsection{Experimental Setup}

\begin{figure*}[t]
    \centering
    \includegraphics[width=0.7\textwidth]{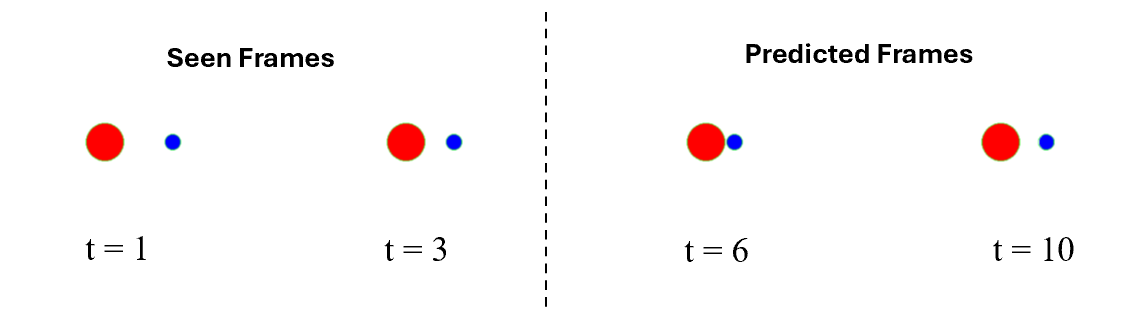}
    \caption{Qualitative visualization of frames predicted by our method for the out of domain, two ball collision case.}
    \label{fig:phyworld_results}
    \vspace{-0.5cm}
\end{figure*}

\begin{table*}[t]
\centering
\resizebox{\textwidth}{!}{
\begin{tabular}{cccccc}
    \toprule
    Model & Training dataset size & uniform-motion-iid ($\downarrow$) & uniform-motion-ood ($\downarrow$) & collision-iid ($\downarrow$) & collision-ood ($\downarrow$) \\
    \midrule 
    DiT-small & 30K & 0.0221 & 0.4349 & 0.0267 & 0.1873 \\
    DiT-big & 30K & 0.0166 & 0.4330 & 0.0302 & 0.2411\\
    DiT-large & 30K & 0.0150 & 0.3783 & 0.0240 & 0.2700\\
    \midrule 
    DiT-small & 3M & 0.0149 & 0.2875 & 0.0227 & 0.1525\\
    DiT-big & 3M & \underline{0.0138} & 0.3583 & \underline{0.0181} & 0.2106 \\
    DiT-large & 3M & \textbf{0.0124} & 0.4270 & \textbf{0.0153} & 0.1613 \\
    \midrule
    Galileo \cite{wu2015galileo} & 30K & 0.0176 & 0.0187 & 0.0502 & 0.0544 \\
    Galileo \cite{wu2015galileo} & 3M & 0.0163 & \underline{0.0173} &  0.0342 & \underline{0.0428} \\
    \midrule
    \textbf{\mname~(Ours)} & 10 & 0.0147 & \textbf{0.0150} & 0.0227 & \textbf{0.0241} \\
    \bottomrule
\end{tabular}
}
\caption{Velocity prediction accuracy for the Phyworld Uniform Ball Motion and Ball Collision settings for both in (iid) and out of distribution (ood) settings. Best results are shown in \textbf{bold}, second best in \underline{underline}.}
\label{tab:phyworld-res}
\end{table*}

\begin{table}[t]
    \centering
    \begin{tabular}{ccccc}
    \toprule 
    \multirow{2}{*}{Model} & \multirow{2}{*}{MAE ($\downarrow$)}  & \multirow{2}{*}{PSNR ($\uparrow$)} & LPIPS ($\downarrow$) & LPIPS ($\downarrow$) \\
    & & & (AlexNet) & (VGGNet) \\
    \midrule 
    TI2V-Zero~\cite{ni2024ti2v} & 0.011 & 22.01 & 5.9e-5 & 0.0038 \\ 
    \midrule
    Stable Video & 0.033 & 18.37 & 2.0e-4 & 0.0046\\
    Diffusion-v1~\cite{blattmann2023stable} \\ 
    \midrule 
    \mname~(Ours) & \textbf{0.003} & \textbf{30.59} & \textbf{1.8e-5} & \textbf{2.8e-4}\\
    \bottomrule
    \end{tabular}
    \caption{Frame prediction performance of competing methods on the Cart Pole environment. Best results shown in \textbf{bold}.}
    \label{tab:cartpole_results}
\end{table}

\begin{figure*}[t]
    \centering
    \includegraphics[width=0.8\textwidth]{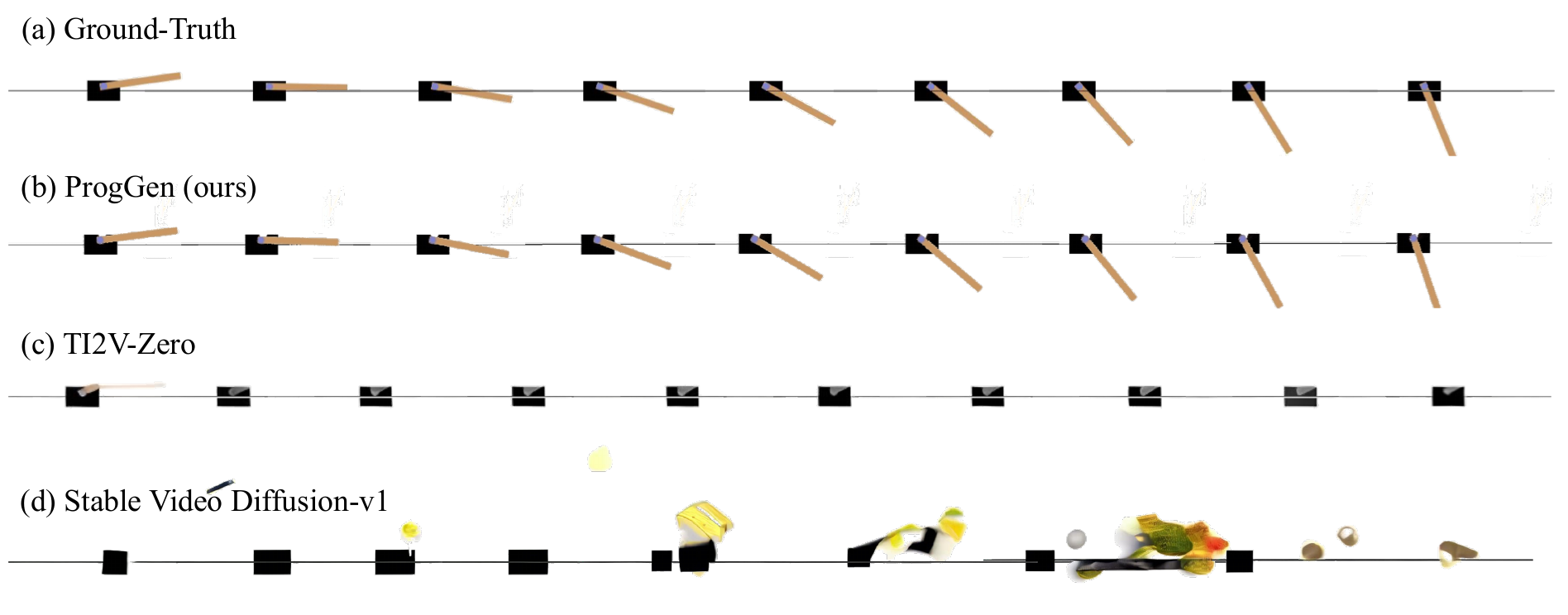}
    \caption{Qualitative visualization of frame prediction results between competing methods on the Cart Pole environment.}
    \label{fig:cartpole_qual}
    \vspace{-0.5cm}
\end{figure*}

\begin{figure}[t]
    \centering
    \includegraphics[width=0.7\linewidth]{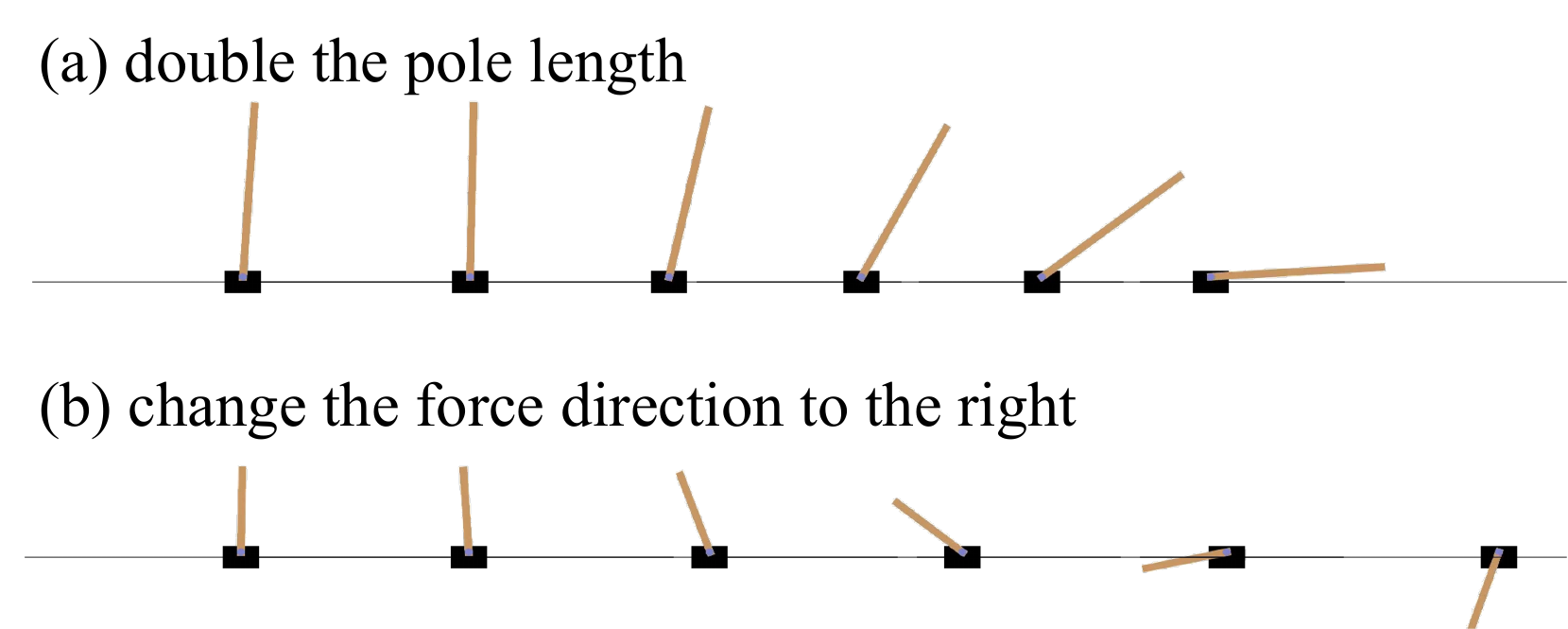}
    \caption{Frame prediction results for our method upon: (a) doubling the pole length; (b) switching the direction of motion the cart.}
    \label{fig:cartpole_counterfactual}
    \vspace{-0.5cm}
\end{figure}


\paragraph{Environments:}
We conduct experiments in the following two challenging simulation environments:

\paragraph{(a) Phyworld-OOD:} The Phyworld environment~\cite{kang2024far} 
simulates the kinematic states for various physical scenarios, such as with balls moving in a confined space and/or colliding with each other. We conduct experiments with videos collected from this environment under two settings: (i) \emph{Uniform:} Here there is only one ball that moves horizontally with a constant velocity in the space. The law of inertia guides the motion of the ball; (ii) \emph{Collision:} Here there are two balls, of different sizes (\ie~ radii), moving horizontally at different velocities and colliding with each other. The law of conservation of energy and momentum guides the collision process and the velocities after the collision. We generate videos under the aforementioned settings by varying the radii and the velocities of the balls. Our programmatic video prediction method works by estimating the positions of the ball to define the states.
The videos contain a total of $20$ frames, with the first $3$ frames being used for conditioning (\ie~ seen frames). To assess the generalizability of competing approaches, we conduct evaluations under two settings, per prior work~\cite{kang2024far}: (i) \emph{In-distribution}: Here the videos in the test set have ball velocities similar to those in the training; (ii) \emph{Out-of-distribution}: Here the test set videos contain ball velocities much higher than those in the training set.
We train all baseline models with varying training set sizes, viz. $30$k and $3$M videos under the aforementioned settings. For \name{}, on the other hand, we use only $10$ videos to train for the parameters, $\theta$. 
\footnote{The code is available at: \href{https://github.com/metro-smiles/ProgGen}{Code Link}}


\paragraph{(b) Cart Pole:} We also conduct experiments in the gym-Cart Pole~\cite{gym} environment, with a constant action -- pulling the cart to the left, while the pole rotates in a clock-wise direction. The videos are generated by altering the pole's initial angle, the velocity of the cart, and the angular velocity of the pole. A programmatic approach to video frame prediction needs to automatically determine that these are the attributes to be estimated and undertake the prediction by extrapolating the dynamics using these attributes. 
In this setting, we test the generalizability of competing approaches by adopting a zero-shot setting where there are no training videos and the models are asked to generate the next ten frames of a video, given the first 10 frames. 

\paragraph{Baselines:}
In order to assess the benefits of \name{} at the task of video frame prediction, we choose the following competitive, state-of-the-art baselines drawn from two broad streams of research in this field: (i) From among the data-driven video frame prediction type of models, we use diffusion-model based methods, such as TI2V-Zero~\cite{ni2024ti2v}, Stable Video Diffusion-v1~\cite{blattmann2023stable} for the CartPole experiments and following Kang \etal~\cite{kang2024far}, use DiT~\cite{kang2024far}  (which follows the same architecture as Sora~\cite{videoworldsimulators2024}) for our experiments on PhyWorld. For DiT, we use the three variants for the PhyWorld dataset, as reported in Kang \etal~\cite{kang2024far}: the DiT-small, DiT-big, and DiT-large, with 22.5M, 89.5M, and 310M parameters, respectively. For TI2V-Zero and Stable Diffusion, we use their pre-trained models available from the web. (ii) From among the physics-based frame prediction approaches, we choose Galileo~\cite{wu2015galileo} for our experiments on PhyWorld. Since the Cart Pole setting does not use any training data, we ignore this baseline for the Cart Pole experiments.

\paragraph{Evaluation Metrics:}

For \underline{Phyworld}, Kang \etal~\cite{kang2024far} uses an error measure based on the estimated velocities of the one or more balls, in every frame, when compared to that used for the synthesis of the videos to assess the quality of prediction. The per-frame velocity is computed as:
\[
 v_t^i = x_{t}^i - x_{t - 1}^i; \forall t \in \{F+1, \dots, T\}, \forall i \in \{1, \dots, B\}
\]
where $v_t^i$ is the velocity of the $i^{th}$ ball in the $t^{th}$ frame, while $x_t^i$ denotes the pixel location of the center of the ball in the $t^{th}$ frame and $B$ is the total number of balls in the environment. The final performance measure is an average of this error over all the balls. For our experiments, we stick to this evaluation measure as well. 


\underline{Cart Pole:} We evaluate the qualities of the generated videos in the Cart Pole environment by comparing them with the ground-truth videos using metrics that assess the quality of reconstructed RGB-frames and reporting the mean performance, including mean-absolute error (MAE), PSNR~\cite{channappayya2008rate}, and LPIPS~\cite{zhang2018unreasonable}. 


\subsection{Results}
\paragraph{PhyWorld:}
As shown in Table~\ref{tab:phyworld-res}, our model significantly outperforms the state-of-the-art diffusion models (\ie DIT-small, DIT-big, DIT-large), trained on $30$k training videos, 
even though our method is only trained on 10 videos. In particular, the velocity error measures for \name{} are an order of magnitude smaller for the out-of-distribution (ood) case when compared to the diffusion-based baseline models. Further, even when the diffusion models are trained with $100\times$ the number of training videos, \ie $3$M, our proposed method still performs comparably. Moreover, we observe that even physics-grounded models such as Galileo~\cite{wu2015galileo}, underperform \name{} under all settings. This is perhaps because of the constraints imposed by Galileo in assuming a Gaussian distribution on the state transition function.
We surmise that the flexibility obtained by modeling the video prediction task programmatically helps in capturing the dynamics of the video at the semantic level, \ie~ in describing the world model in terms of objects and their motion rather than in terms of changes at the pixel level. Additionally, the global parameters $\theta$ accommodate for variance within the type of motion structure captured in the programs allowing \name{} to fit to a wide range of videos.

Figure~\ref{fig:phyworld_results} visualizes the frames generated by our method for the two ball, out-of-distribution collision case. 
We see that our method is successful at following the physical laws as specified by the programs and can generate visually plausible frames. 
We also provide ablation results for training the DIT models with $300$k training videos in the appendix.



\paragraph{Cart Pole:}
To assess the generalizability of \name{}, we evaluate our method against competing baselines on the gym-Cart Pole environment.
As shown in Table~\ref{tab:cartpole_results}, our method outperforms all competing baselines, across all measures, by significant margins in this environment. This is evident from the qualitative results in Figure~\ref{fig:cartpole_qual}, as well. Furthermore, as shown in Figure~\ref{fig:cartpole_counterfactual}, \name{} can accommodate counter-factual reasoning effectively, such as if the pole length were to be doubled or the direction of motion of the cart were to be reversed, and still generate plausible frames. 




%
\section{Conclusions}
\label{sec:conclusion}

In this work, we propose \name{}, a program synthetic approach to video frame prediction which relies on the inductive biases of LLM/VLM to construct world models of videos, represented as a sequence of neuro-symbolic, human-interpretable states. Our proposed approach, while still largely modeling the video in an object-centric fashion, can outperform state of the art diffusion models for the task across multiple challenging environments, while utilizing only a fraction of the training data. 

\noindent \textbf{Limitations:} Adapting \name{} to work in more real world settings might require a richer set of attribute descriptions which might be challenging for current LLM/VLM to estimate.




\bibliography{example_paper}
\bibliographystyle{iclr2025_conference}

\clearpage
\section{Appendix}
\label{sec:appendix}




\subsection{Model Performance with Varying Training Set Sizes}

\begin{table*}[h]
\centering
\resizebox{\textwidth}{!}{
\begin{tabular}{cccccc}
    \toprule
    Model & Training dataset size & uniform-motion-iid ($\downarrow$) & uniform-motion-ood ($\downarrow$) & collision-iid ($\downarrow$) & collision-ood ($\downarrow$) \\
    \midrule 
    DiT-small & 30K & 0.0221 & 0.4349 & 0.0267 & 0.1873 \\
    DiT-big & 30K & 0.0166 & 0.4330 & 0.0302 & 0.2411\\
    DiT-large & 30K & 0.0150 & 0.3783 & 0.0240 & 0.2700\\
    \midrule 
    DiT-small & 300K & 0.0184 & 0.2929 & 0.0241 & 0.2000\\
    DiT-big & 300K & 0.0166 & 0.4330 & 0.0193 & 0.1663\\
    DiT-large & 300K & 0.0140 & 0.3847 & 0.0158 & 0.1725 \\
    \midrule 
    DiT-small & 3M & 0.0149 & 0.2875 & 0.0227 & 0.1525\\
    DiT-big & 3M & 0.0138 & 0.3583 & 0.0181 & 0.2106 \\
    DiT-large & 3M & 0.0124 & 0.4270 & 0.0153 & 0.1613 \\
    \midrule 
    Galileo \cite{wu2015galileo} & 30K & 0.0176 & 0.0187 & 0.0502 & 0.0544 \\
    Galileo \cite{wu2015galileo} & 3M & 0.0163 & 0.0173 &  0.0342 & 0.0428 \\
    \midrule
    Ours & 1 & 0.0192 & 0.0185 & 0.4805 & 0.2879 \\
    Ours & 10 & 0.0187 & 0.0176 & 0.0385 & 0.0418\\
    Ours & 100 & 0.0175 & 0.0164 & 0.0340 & 0.0321 \\
    \bottomrule
\end{tabular}
}
\caption{Velocity prediction accuracy for the Phyworld Uniform Ball Motion and Ball Collision settings for both in (iid) and out of distribution (ood) settings. } 
\label{tab:phyworld_full}
\end{table*}

In Table~\ref{tab:phyworld_full}, we report model performances with varying training set sizes for the PhyWorld~\cite{kang2024far} environment. We train the competing baseline models with $30$k, $300$k, or $3$M videos, while we train our proposed approach with merely $1$, $10$ or $100$ training videos. As we see from the results, our proposed method (\name{}) trained with $10$ or $100$ videos, outperforms all competing methods for the challenging, out of the distribution (ood) motion setting. Noticeably, our model trained with just $1$ video is also competitive in the ood setting. Even for the in distribution setting (iid), our model is competitive with respect to the competing baselines.

\subsection{Illustrative Prompts and Codes}

Below, we present as an illustrative example of the prompt used in order for \name{} to generate the \emph{Dynamics Estimation Program} $\mathcal{D}$, using a VLM, for the Cart Pole environment: 

\textcolor{blue}{``The environment you're in is the classic CartPole environment. This is a well-known problem in reinforcement learning where a pole is attached by an un-actuated joint to a cart, which moves along a frictionless track. The goal is to keep the pole upright by moving the cart left or right. The state of the system is represented by a tuple consisting of: 1. cart\_position: The position of the cart on the track. 2. cart\_velocity: The velocity of the cart. 3. pole\_angle: The angle of the pole with the vertical axis. 4. pole\_angular\_velocity: The angular velocity of the pole. Given this setup, the dynamics of the environment are governed by the physics of the system, specifically the equations of motion for the cart and the pole. The force applied to the cart (left or right) will affect these state variables over time. We'll use some constants (which will be optimized later) such as gravity, mass of the cart and pole, length of the pole, and the time step for the discretization of the dynamics.''}

The above prompt leads to the following synthesized program:

\small
\begin{lstlisting}
def transition(state: jnp.ndarray, params: jnp.ndarray):
    x, x_dot, theta, theta_dot = state
    mass_cart, mass_pole, length, force = params[:4]
    total_mass = mass_cart + mass_pole
    cos_theta, sin_theta = jnp.cos(theta), jnp.sin(theta)
    temp = (force + pole_mass_length * theta_dot ** 2 * sin_theta) / total_mass
    theta_acc = (gravity * sin_theta - cos_theta * temp) / 
                        (length * (4.0/3.0 - mass_pole * cos_theta ** 2 / total_mass))
    x_acc = temp - pole_mass_length * theta_acc * cos_theta / total_mass
    x = x + time_step * x_dot
    x_dot = x_dot + time_step * x_acc
    theta = theta + time_step * theta_dot
    theta_dot = theta_dot + time_step * theta_acc
    return jnp.array([x, x_dot, theta, theta_dot])
\end{lstlisting}
\normalsize

\subsection{Failure Cases}


\begin{figure}
    \centering
    \includegraphics[width=0.3\linewidth]{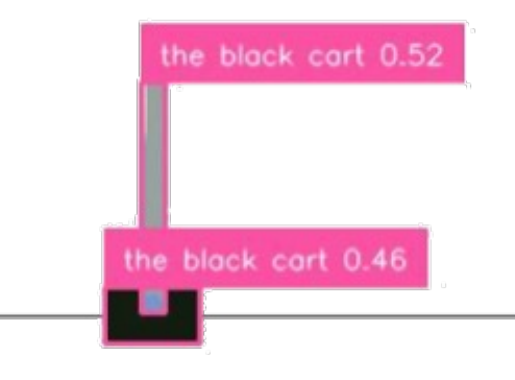}
    \caption{Grounded-SAM~\cite{ren2024grounded} incorrectly labels the pole as "the black cart". }
    \label{fig:sam_fail}
\end{figure}

Even though \name{} achieves state-of-the-art frame prediction performance in our experiments, it does falter sometimes. In Figure~\ref{fig:sam_fail}, we show one such case. Here, the detections produced by Grounded SAM~\cite{ren2024grounded} are incorrect. In particular, even the ``pole'' is detected as a ``black cart'' with high confidence. In order to correct for such erroneous detections, we leverage GPT-4v~\cite{achiam2023gpt} to verify the detections in order to improve the effectiveness and reliability of the perception but we observe GPT-4v's perception module to also be noisy at times resulting in inaccurate estimation of attributes, for such cases.

\end{document}